\documentclass[10pt,english]{article}
\usepackage{subfigure}
\usepackage{graphicx}
\usepackage{float}
\usepackage[T1]{fontenc}
\usepackage[latin9]{inputenc}
\usepackage[margin=1.5in]{geometry}
\usepackage{float}
\usepackage{amsmath}
\usepackage{amsbsy}
\usepackage{setspace}
\usepackage{amssymb}
\usepackage{esint}
\usepackage{cite}
\newfloat{algorithm}{tbp}{loa}
\floatname{algorithm}{Algorithm}
\usepackage{algorithmic}
\usepackage{hyperref}

\newlength\myindent
\makeatletter

\floatstyle{ruled}
\newfloat{algorithm}{tbp}{loa}
\floatname{algorithm}{Algorithm}

\usepackage{babel}

\begin{document}

\title{Autoencoders as Pattern Filters}

\author{M. Andrecut}


\maketitle
{

\centering Calgary, Alberta, Canada

\centering mircea.andrecut@gmail.com

} 

\bigskip 

\begin{abstract}

We discuss a simple approach to transform autoencoders into "pattern filters". 
Besides filtering, we show how this simple approach can be used also to build robust classifiers, 
by learning to filter only patterns of a given class. 

\bigskip


\end{abstract}

\bigskip

\section{Introduction}

An autoencoder is a feed-forward Deep Neural Network (DNN) consisting of an encoder and a decoder, trained to reproduce its input at the output layer (Figure 1) \cite{key-1}, \cite{key-2}. 
Autoencoders are used in a large variety of applications: dimensionality reduction, feature extraction, image denoising, imputing missing data etc.
Depending on the dimensionality of the hidden layer we can distinguish two types of autoencoders:
\begin{itemize}
\item Undercomplete: the hidden layer has a lower dimension than the input/output layers. 
\item Overcomplete: the hidden layer has a higher dimension than the input/output layers. 
\end{itemize}
In general, the dimensionality of the hidden layer should be different than the dimensionality of the input/output layers in order to avoid learning an identity data transformation. 

Undercomplete autoencoders are typically used in unsupervised learning tasks, such as: dimensionality reduction, feature learning, and generative models.
The encoder is generally used to learn a lower dimensional latent representation of the input samples, performing an efficient compression through non-linear transformations. 
In the same time, the decoder learns how to reconstruct the input samples from this latent compressed representation. 

Due to the higher dimension of the hidden layer, overcomplete autoencoders are prone to copy the input to the output rather than learning important features. 
This is why overcomplete autoencoders require some form of regularization, such as sparsity constraints, for practical applications. 

\begin{figure}[!h]
\centering \includegraphics[width=5cm]{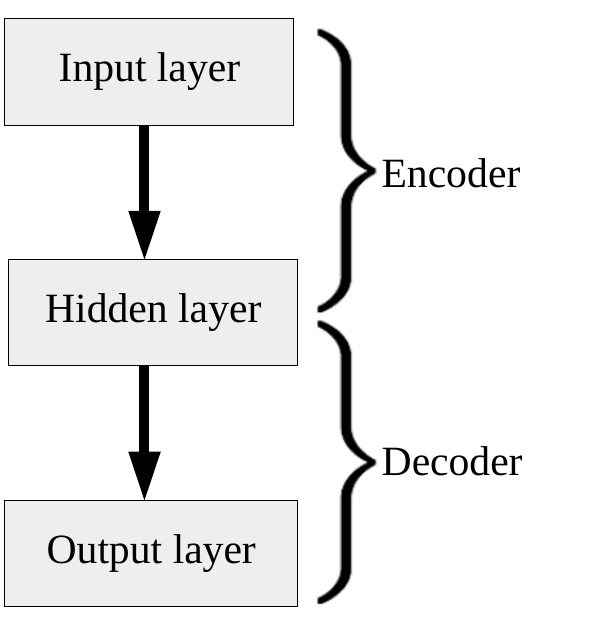}
\caption{Autoencoder schema.}
\end{figure}

Here we show how an autoencoder can be thought to learn a single distinct class of patterns, such that it can perform an efficient "pattern filtering". We may think about such "pattern filters" in an analogy to their 
optical counterparts. However, optical filters are used only to filter certain frequencies of the light spectrum, while here we would like to filter the patterns corresponding to distinct input classes of samples. 
We show that besides filtering, this approach can be used also to build robust classifiers, by learning to filter only patterns of a given class. 
That is, for each distinct class we train an autoencoder which is becoming "transparent" to the samples extracted from this particular class, and in the same time it becomes "opaque" to the samples extracted from the other classes. 
This way, we end up with as many autoencoders as the classes considered, however these autoencoders can then be used very easily to build a quite robust pattern classifier. 

\section{Pattern filtering with autoencoders}

Let us assume that the input (output) space is $\mathbb{R}^n$, and the encoded space is $\mathbb{R}^m$. Using the encoder, an input sample $x\in \mathbb{R}^n$ is encoded into:
\begin{equation}
E(w_E,x)=\tilde{x}\in \mathbb{R}^m,
\end{equation}
and then reconstructed using the decoder into:
\begin{equation}
D(w_D,\tilde{x})=\hat{x}\in \mathbb{R}^n.
\end{equation}
Thus, the encoder is a non-linear mapping transformation from the input space to the encoded space, 
$E:\mathbb{R}^n\rightarrow\mathbb{R}^m$, while the decoder is a "reciprocal" transformation from the encoded space into the output space (which is the same as the input space), $D:\mathbb{R}^m\rightarrow\mathbb{R}^n$. 
The learning consists in finding the encoding and decoding parameters (DNN weights), such that the autoencoder minimizes the distance between the input $x$ and the output $\hat{x}$:
\begin{equation}
w^*_E,w^*_D = \min_{w_E,w_D} L(x,D(w_D,E(w_E,x))) = \min_{w_E,w_D} \Vert x - D(w_D,E(w_E,x)) \Vert^2_2, 
\end{equation}
where $\Vert.\Vert$ is the $\ell_2$ norm (Euclidean).
This learning task can be accomplished using DNN techniques such as the stochastic gradient implementation of the back propagation algorithm. 

In the case of undercomplete autoencoders, the dimensionality of the encoded space is set to a smaller value than the dimensionality of the input/output space, 
$m<n$, in order to prevent the autoencoder from learning the identity function, and to force it to learn a richer latent representation. 
On the other hand, the overcomplete autoencoder avoids learning a trivial identity function by imposing sparsity restrictions on the hidden weights of the autoencoder:
\begin{equation}
w^*_E,w^*_D = \min_{w_E,w_D} L(x,D(w_D,E(w_E,x))) + \alpha L_{s}(w_E,w_D), 
\end{equation}
where $\alpha > 0$ measures the sparsity weight in the minimization process. The typical loss regularization function $L_s$ is the $\ell_1$ norm: $\Vert w \Vert_1$. 

We should note that "by design", an autoencoder is an auto-associative DNN, which means that a training input sample $x$ is associated with itself: $A(x)\simeq x$. 
In order to solve the pattern filtering and classification problems here we "slightly" deviate from the traditional autoencoder approach, and we would like to consider a more relaxed "associative autoencoder" requirement, 
such that a training input sample $x$ is not necessarily associated with itself, but with another class of patterns, which is completely unrelated with any of the considered classes. 

Let us assume that we have $K$ distinct input classes of patterns:
\begin{equation}
C_k = \lbrace x_i^{(k)} \vert x_i^{(k)}\in \mathbb{R}^n, i=0,1,...,N^{(k)} \rbrace,\quad k=0,1,...,K-1,
\end{equation}
where $N^{(k)}$ are the number of samples in the class $k$.
For each class $k$ we wish to build an autoencoder $A_k$ which is "transparent" to the input samples from this class, and "opaque" to the samples from all other classes $j\neq k$.
Then, using these $K$ autoencoders, an unknown sample $x$ can be classified as $x \in C_k$ if:
\begin{equation}
\Vert x - A_k(x) \Vert_2 < \Vert x - A_j(x) \Vert_2, \quad \forall j\neq k.
\end{equation}

\section{Transparency and opacity in autoencoders}

Transparency is an "intrinsic" property of the autoencoder, since its purpose is to minimize the dissimilarities between the input and output patterns. 
The remaining question is how to make an autoencoder "selective", that is "transparent" for a given class and "opaque" to all other classes? 
One way to do this is to train the autoencoder $A_k$ such that the input samples from the target class $C_k$ are associated with themselves in the output, $x_i^{(k)} \leftrightarrow x_i^{(k)}$, while 
all other training samples drawn from different classes $C_j$, $j\neq k$, are associated with randomly (noisy) generated patterns taking values in the same domain, $x_i^{(j)} \leftrightarrow \xi_i^{(j)}$. 
Hopefully this will map the $C_k$ samples to themselves, and the $C_j$, $j\neq k$, samples to random noise, and therefore the autoencoder becomes "selective". 
This is the approach we adopt here, but there may be also other possible options.

\section{Numerical implementation}

In order to illustrate the "filtering" and classification abilities of the "associative autoencoders" here we use two popular data sets: MNIST \cite{key-3} and fashion-MNIST \cite{key-4}. 
The MNIST data set is a large database of handwritten digits $\{0,1,...,9\}$, containing 60,000 training images 
and 10,000 testing images. These are monochrome images with an intensity in the interval $[0,255]$, and the size of $28 \times 28 = 784$ pixels. 
The MNIST data set is probably the most frequently used benchmark in image classification. 
The fashion-MNIST dataset also consists of 60,000 training images and a test set of 10,000 images. The images are also monochrome, with an intensity in the interval $[0,255]$ and the size of $28 \times 28 = 784$ pixels. 
However, the fashion-MNIST is a more complex dataset that contains images from $K=10$ different apparel classes: 
0 - t-shirt/top; 1 - trouser; 2 - pullover; 3 - dress; 4 - coat; 5 - sandal; 6 - shirt; 7 - sneaker; 8 - bag; 9 - ankle boot. 

The simplest autoencoder consists of an input layer of neurons with the same dimension as the input samples, followed by a hidden layer of neurons (the encoder, which can be undercomplete or overcomplete), and an output layer, 
containing the same number of neurons as the input layer (the decoder):
\begin{equation}
E_k(w_E,x) = f(w_E x+a_E) = \tilde{x}, D_k(w_D,\tilde{x})= g(w_D x + b_D)= \hat{x}. 
\end{equation}
Here, $f$ and $g$ are the neuron activation functions, and $a_E \in \mathbb{R}^m$ and $b_D  \in \mathbb{R}^n$ are the bias parameters (to be learned also, typically they are included in the weights sets in the 
optimization problems). The performance of this basic autoencoder can be improved by including more intermediate layers into the encoder and decoder.  
However, here we only consider the basic autoencoder, since the goal of the paper is only to provide a simple proof of concept, and show that the idea is feasible. 

Thus, the basic "associative autoencoder" considered here consists of three neuron layers with the following properties:
\begin{itemize}
\item Layer 1: the input layer of neurons with the same dimensionality as the input samples, $n=784$, and no activation function;
\item Layer 2: the hidden layer with a dimensionality $n^*$, and a "relu" activation function;
\item Layer 3: the output layer with dimensionality $n$, and a "sigmoidal" activation function;
\end{itemize}

We trained the autoencoder using the Python Keras library \cite{key-5}, acting as an interface for the TensorFlow library \cite{key-6}. Both MNIST and fashion-MNIST datasets are included in Keras package. 
We used the "mean squared error" loss function and the "adam" optimizer, for 125 epochs and a batch size of 250 images. 
Also, for the overcomplete case, when $n^* \geq n$, we used the $\ell_1$ sparsity regularization constraint, by applying the Keras $\ell_1$ weight regularizer to the hidden layer. 
We should note here that the regularization is not really necessary for the overcomplete case, since the "associative autoencoder" is more relaxed and it is not tasked with reproducing its input, 
and therefore it cannot learn an identity mapping as in the case of a "strict" autoencoder. The role of the sparsity regularization used here is mostly to obtain a more simplified and compact model after learning. 

\section{Numerical results}

The accuracy results for the basic "associative autoencoder" are shown in Figure 2 for the MNIST data set, and respectively in Figure 3 for the fashion-MNIST data set. 
Here we let the dimensionality of the hidden layer to vary as $n^*=2^\ell$, where $\ell \in \{0,1,...,12\}$, and we measured the accuracy $\eta$ as a function of $n^*$. 

\begin{figure}[!ht]
\centering \includegraphics[width=11.75cm]{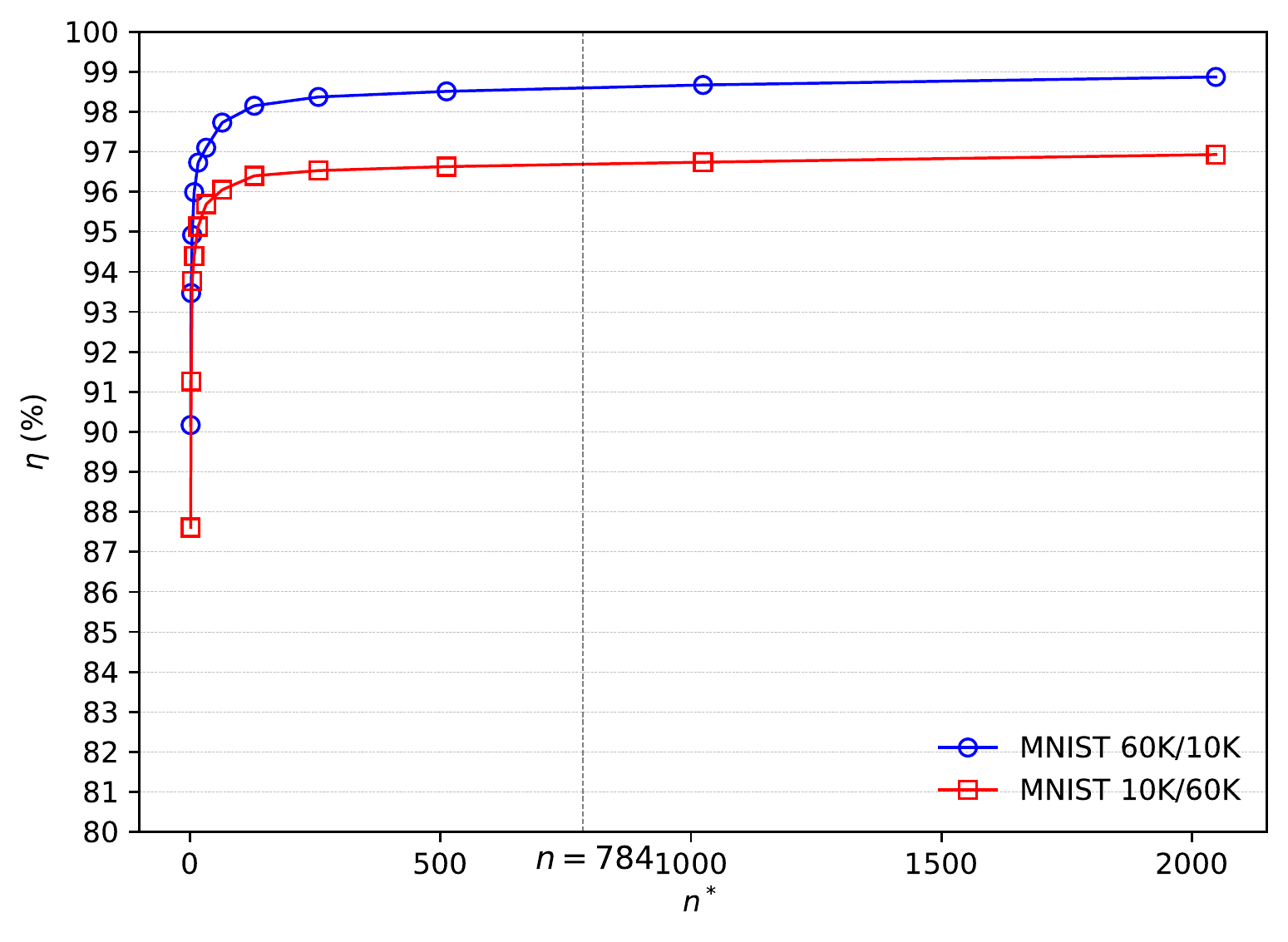} 
\caption{Classification accuracy for MNIST.}
\end{figure}

\begin{figure}[!ht]
\centering \includegraphics[width=11.75cm]{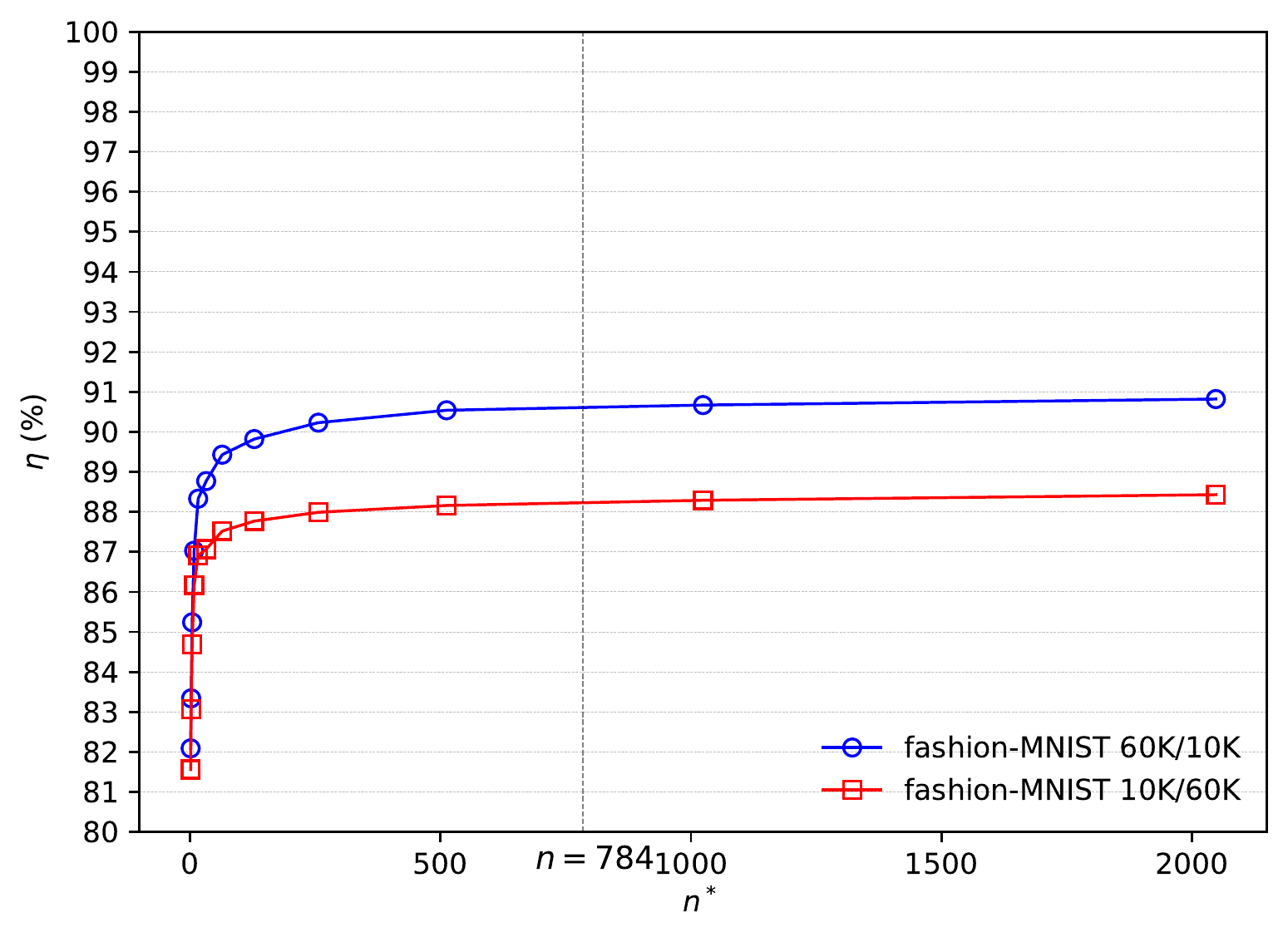}
\caption{Classification accuracy for fashion-MNIST.}
\end{figure}

\begin{figure}[!ht]
\centering \includegraphics[width=8.15cm]{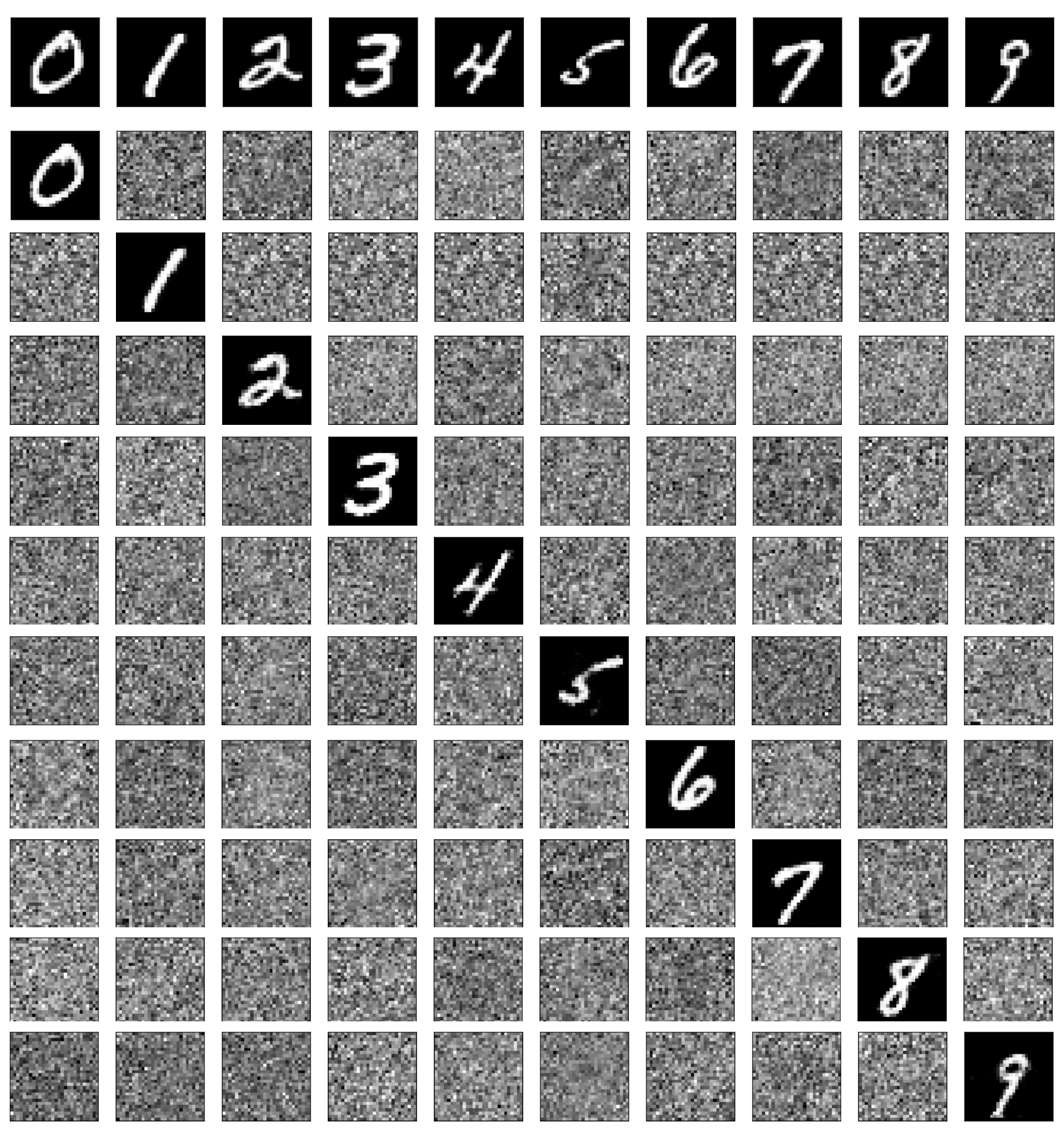} 
\caption{Filtering ability for MNIST.}
\end{figure}

\begin{figure}[!ht]
\centering \includegraphics[width=8.15cm]{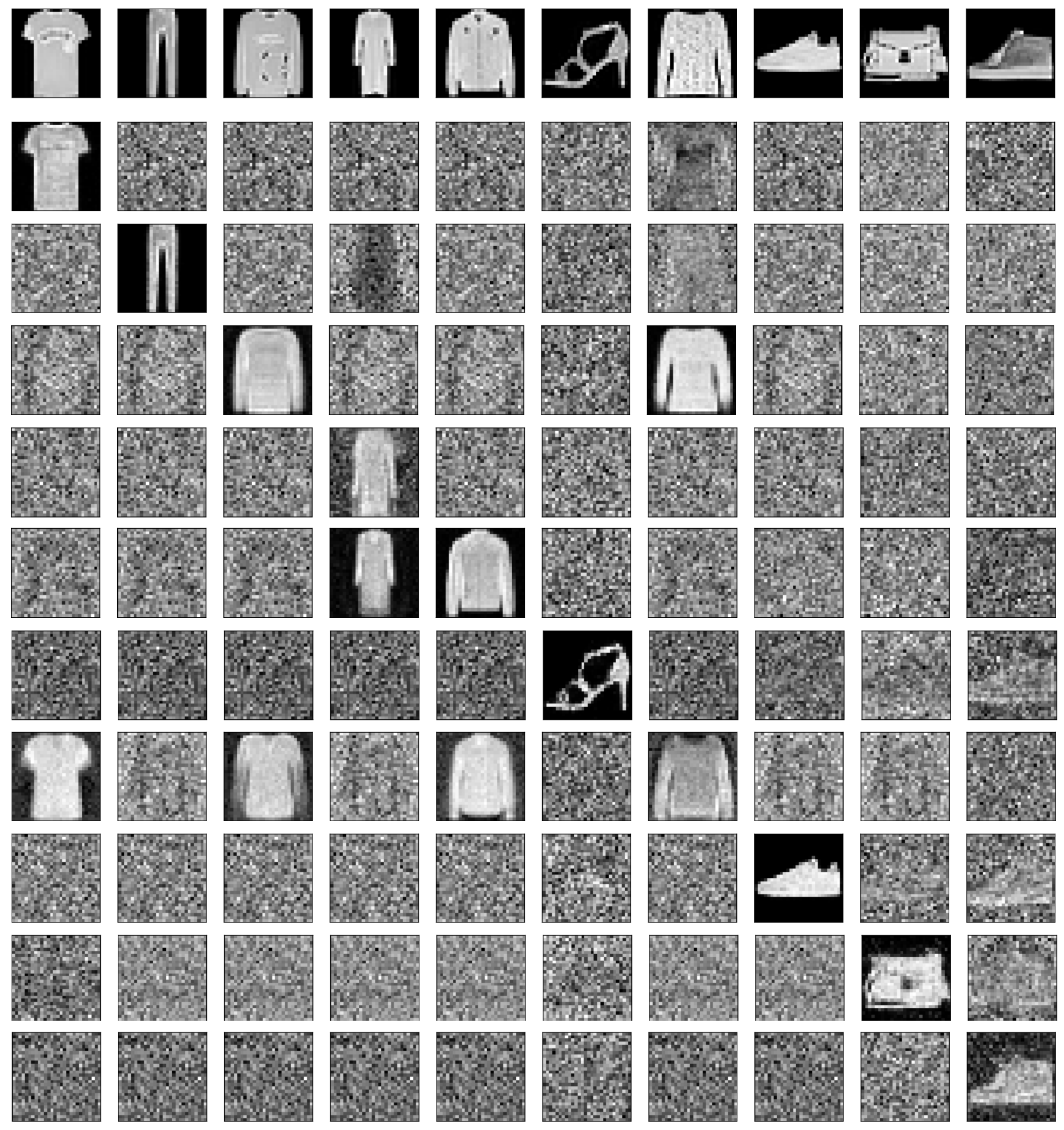} 
\caption{Filtering ability for fashion-MNIST.}
\end{figure}

One can see that for both data sets the accuracy increases very fast in the "undercomplete regime", where $n^*<n$, and it becomes almost stationary in the "overcomplete regime". 
In the case of MNIST, the accuracy increases quickly to $98\% < \eta < 99\%$ for $n\geq 2^7=128$. 
The fashion-MNIST is much harder to classify, but for $n\geq 2^8=256$ the obtained accuracy is $90\% < \eta < 91\%$. 
These results are quite good considering the simplicity of the method. 
The most striking result is that even for a very low encoding dimension, like $n^*=2$, the classification accuracy is $\eta>93\%$ for MNIST, and respectively $\eta>83\%$ for fashion-MNIST. 
These results show that the "asssociative autoencoders" are quite robust with respect to the size of the encoding dimension. 
The "robustness" of the method can also be tested by switching the training and testing sets among them. 
Initially we trained on 60,000 samples and we tested on 10,000 samples (the 60K/10K cases in the figures), but now we switch to 
training only on 10,000 samples, and testing on 60,000 samples (the 10K/60K cases in the figures). 
One can see that after switching the accuracy drops just by $\sim2\%$, maintaining a high value even for lower encoding dimensions. 

In Figure 4 and Figure 5 we also illustrate how the "filtering" performs for each "associative autoencoder" class. In this particular case we assumed $n^*=512$. 
The first row in each figure contains a sample from each class (input samples), the subsequent 10 rows correspond to the output of the "associative autoencoder" of each class, from 0 (top) to 9 (bottom). 
For the MNIST data set, one can see that for each class, the corresponding autoencoder is "transparent" for the sample extracted from the same class, and "opaque" for the samples extracted from different classes, 
which explains the high classification accuracy values. 
In the case of fashion-MNIST, there is a strong interference of class 6 with the classes 0, 2, and 4, explaining the lower classification accuracy values. 
This is because there is a strong similarity in the images of the "shirt" (class 6) and "t-shirt" (class 0), "pullover" (class 2), and "coat" (class 4). Otherwise, the filtering performs quite well. 

\section*{Conclusion}

We have shown how an "associative autoencoder" can learn a single distinct class of patterns, such that it can perform efficient "pattern filtering". Also we have shown how to use 
this "filtering" property to build robust classifiers.  
In this approach, for each distinct class we train an "associative autoencoder" such that it becomes "transparent" to the samples extracted from a particular class, 
and in the same time it becomes "opaque" to the samples extracted from the other classes. The numerical results obtained with a basic three layer "associative autoencoder", for the MNIST and fashion-MNIST data sets, 
show quite good results considering the simplicity of the method. The results can be improved by including more intermediate layers, and creating a "deeper" neural network. 

\section*{Appendix}

For reproducibility reasons we provide the minimal code necessary to implement the "associative autoencoder" using the Keras and TensorFlow libraries. 
One can select between MNIST and fashion-MNIST, 60K/10K and 10K/60K scenarios, by commenting and uncommenting the corresponding lines in the code given below:

\footnotesize
\begin{verbatim}
import numpy as np
import tensorflow as tf
from tensorflow import keras
from tensorflow.keras import layers
from tensorflow.keras import regularizers

if __name__ == "__main__":
#   MNIST data
    (x_train, y_train), (x_test, y_test) = keras.datasets.mnist.load_data() # 60K/10K
#   (x_test, y_test), (x_train, y_train) = keras.datasets.mnist.load_data() # 10K/60K
#   fashion-MNIST data
#   (x_train, y_train), (x_test, y_test) = keras.datasets.fashion_mnist.load_data() # 60K/10K
#   (x_test, y_test), (x_train, y_train) = keras.datasets.fashion_mnist.load_data() # 10K/60K

#   Reshape images
    (N,L,L),M = np.shape(x_train),len(x_test)
    x_train = np.reshape(x_train.astype('float32')/255,(N,L*L))
    x_test = np.reshape(x_test.astype('float32')/255,(M,L*L))

    K,(N,J),M,models = np.max(y_train)+1,np.shape(x_train),len(x_test),[]
    L = 256 # encoding dimmension (hidden layer)
    regularization = False
    for k in range(K):
        print("k=",k)
        x = x_train[y_train==k]
        z = x_train[y_train!=k]
        r = np.random.rand(len(z),J)
        model = keras.Sequential()
        model.add(layers.Input(shape=(J,)))
        if regularization:
            model.add(layers.Dense(L, activity_regularizer=regularizers.l1(1e-5), activation="relu"))
        else:
            model.add(layers.Dense(L, activation="relu"))
        model.add(layers.Dense(J, activation="sigmoid"))
        model.compile(loss="mean_squared_error",optimizer="adam")
        model.fit(np.vstack((x,z)),np.vstack((x,r)),verbose=0,epochs=125,batch_size=250,shuffle=True)
        models.append(model)

#   Classification
    d = np.zeros((M,K))
    for k in range(K):
        xx = models[k].predict(x_test)
        d[:,k] = np.linalg.norm(xx-x_test,axis=1)
    a = 0 # accuracy
    for m in range(M):
        i = np.argmin(d[m,:])
        if i == y_test[m]:
            a += 1
    print("accuracy=",np.round(a*100./M,3),"%")
\end{verbatim}

\end{document}